\documentclass[conference]{IEEEtran}
\IEEEoverridecommandlockouts

\usepackage{cite}
\usepackage{amsmath,amssymb,amsfonts}
\usepackage{algorithmic}
\usepackage{graphicx}
\usepackage{textcomp}
\usepackage{xcolor}
\usepackage{booktabs}
\usepackage{multirow}
\usepackage{bm}
\def\BibTeX{{\rm B\kern-.05em{\sc i\kern-.025em b}\kern-.08em
    T\kern-.1667em\lower.7ex\hbox{E}\kern-.125emX}}
\begin{document}

\title{SpikeDS: Dual Sparsity Spikformer for Perineural Invasion Prediction in 3D MRI}

\author{
\IEEEauthorblockN{
Induk Um$^{1}$, Youngung Han$^{2}$, Kyeonghun Kim$^{3}$, Yului Jeong$^{2}$, Jina Jeong$^{2}$, Hyunsu Go$^{2}$,\\
Dohyun Kweon$^{4}$, Sungha Park$^{2,5}$, Junga Kim$^{2}$, Anna Jung$^{2}$, Suah Park$^{2}$,\\
 Hyuk-Jae Lee$^{2}$, Pa Hong$^{6}$, Woo Kyoung Jeong$^{7}$, Won Jae Lee$^{6}$, Ken Ying-Kai Liao$^{8}$, Nam-Joon Kim$^{2,{\dag}}$
}

\IEEEauthorblockA{
$^{1}$Chung-Ang University, Seoul, Republic of Korea \quad
$^{2}$Seoul National University, Seoul, Republic of Korea \quad
}
\IEEEauthorblockA{
$^{3}$OUTTA,  Seoul, Republic of Korea \quad
$^{4}$Kyung Hee University, Seoul, Republic of Korea}

\IEEEauthorblockA{
$^{5}$Seoul National University School of Medicine, Seoul, Republic of Korea
}

\IEEEauthorblockA{
$^{6}$Samsung Changwon Hospital, Changwon, Republic of Korea\quad
$^{7}$Samsung Medical Center, Seoul, Republic of Korea
}
\IEEEauthorblockA{
$^{8}$NVIDIA AI Technology Center, Taipei, Taiwan}

\IEEEauthorblockA{
$^{\dag}$Corresponding authors:  
\texttt{knj01@snu.ac.kr}
}
}

\maketitle

\begin{abstract}
Perineural invasion (PNI) is associated with poor prognosis in cholangiocarcinoma (CCA). However, its detection from 3D MRI remains challenging due to the subtle and spatially heterogeneous imaging signatures at the tumor periphery. Capturing such spatially sparse cues necessitates volumetric analysis of 3D MRI, but existing deep learning
approaches incur prohibitive computational costs on volumetric
medical images, limiting their clinical deployment.
We propose Dual Sparsity Spikformer (SpikeDS), a spiking neural network architecture that jointly exploits \emph{activation sparsity} from binary spike communication and \emph{spatial sparsity} from window pruning on the firing rate.
SpikeDS introduces Dual Sparsity Spiking Attention (DSSA), which combines two complementary mechanisms. The first is Window-based Expert Mixture Spiking Attention (W-EMSA), which selectively applies attention only to salient windows identified by their firing rates; the second is Cross-Window Spiking Self-Attention (CW-SSA), which enables global context exchange through an asymmetric scheme where pruned windows still contribute as key–value sources.
Evaluated on a clinical cohort of 139 CCA patients via 5-fold 
cross-validation, SpikeDS achieves an AUC of 0.753 at only 14.4\,mJ---surpassing the best baseline in both AUC and energy efficiency. These results suggest that dual sparsity provides an effective hardware-aware strategy for improving the efficiency of 3D spiking transformers without compromising diagnostic performance.

\end{abstract}


\section{Introduction}


Perineural invasion (PNI) refers to the infiltration of tumour cells into the space surrounding peripheral nerves, constituting a distinct pathway for tumour spread beyond the primary site~\cite{liebig2009perineural}.
In cholangiocarcinoma (CCA), PNI has been closely linked to higher recurrence rates, lymph node metastasis, and diminished overall survival, establishing it as a prognostic factor following surgical intervention~\cite{conti2026perineural,li2020perineural,wei2022prognostic}.
Preoperative identification of PNI is clinically valuable for guiding the extent of surgical resection~\cite{chen2025preoperative} and
neoadjuvant treatment decisions~\cite{meng2023clinical}; however, PNI manifests as microscopic neural tracking at the tumour 
periphery~\cite{liebig2009perineural}, making non-invasive detection 
inherently difficult~\cite{han2026losanet,han2026mmaformer,han2026neonet}.
Conventional quantitative approaches have predominantly employed radiomics-based analysis~\cite{huang2021feasibility}, which demands extensive manual delineation of tumour boundaries~\cite{larue2017quantitative,lambin2017radiomics}.

Recent advances in 3D deep learning have enabled automated volumetric analysis for medical imaging.
CNN-based architectures such as ResNet~\cite{he2016deep}, DenseNet~\cite{huang2017densely}, and EfficientNet~\cite{tan2019efficientnet} have been widely adopted for 3D classification tasks, while Transformer-based models like Swin~Transformer~\cite{liu2021swin} leverage window-based self-attention to capture long-range dependencies.
However, applying self-attention to high-resolution 3D volumes incurs a computational burden that scales quadratically with the number of tokens, making it impractical for resource-constrained clinical settings.

Spiking Neural Networks (SNNs) offer a fundamentally different computational paradigm that addresses this energy bottleneck.
Unlike conventional Artificial Neural Networks (ANNs) that rely on energy-intensive floating-point multiply-accumulate (MAC) operations, SNNs communicate via asynchronous binary spikes; when activations are zero, the corresponding multiplications can be entirely skipped, reducing both operations and energy~\cite{roy2019towards,yin2021accurate}.
Spikformer~\cite{zhou2022spikformer} first demonstrated that spiking self-attention (SSA), using spike-form queries, keys, and values without softmax, can achieve competitive performance on vision tasks.
The Spike-driven Transformer (SD-Transformer)~\cite{yao2023spike} further advanced this line by proposing a purely spike-driven attention mechanism, and its successor Meta-SpikeFormer~\cite{yao2024spike} established a versatile meta-architecture for SNNs.
More recently, the Spiking Transformer with Experts Mixture (SEMM)~\cite{zhou2024spiking} introduced a binary routing mechanism that selectively activates expert heads through Hadamard masking, realising sparse, event-driven mixture-of-experts routing entirely within the spike domain.

\begin{figure*}[t]
\centering
\includegraphics[width=1\textwidth]{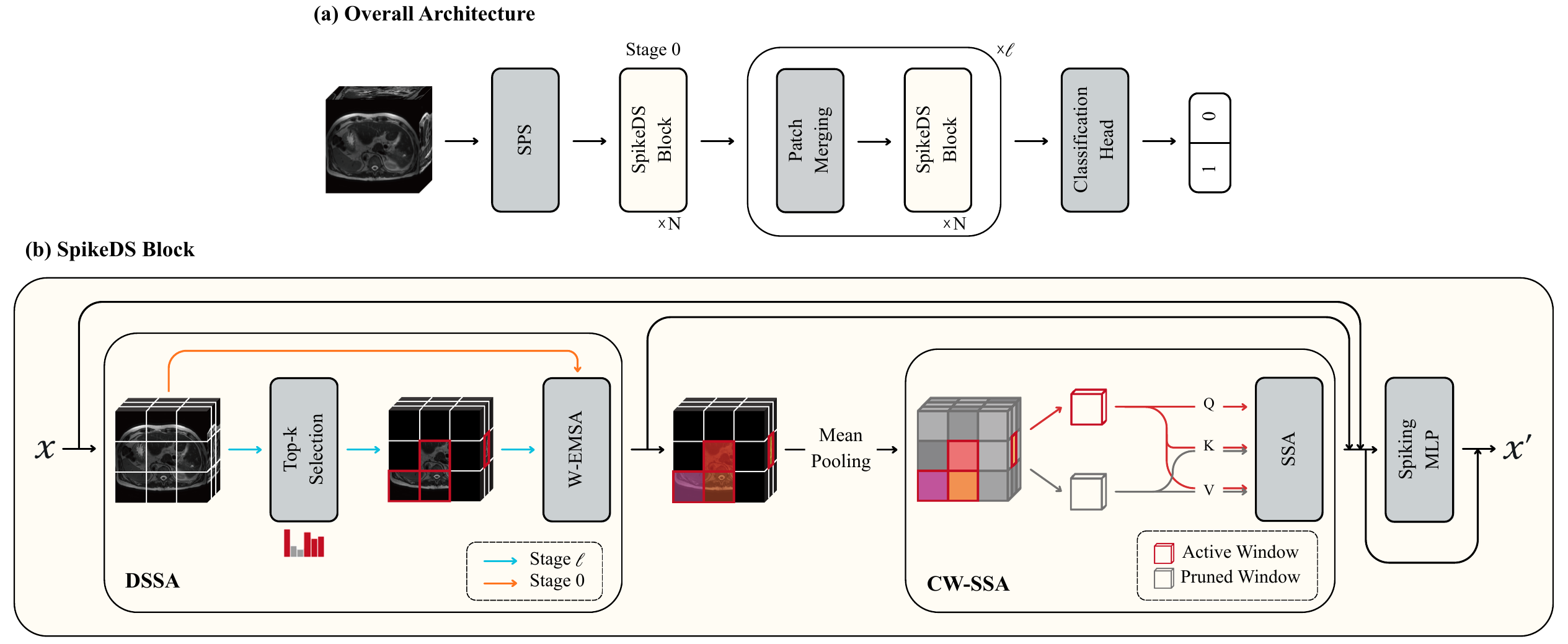}
\caption{(a) Overall architecture of SpikeDS. A 3D MRI volume is converted into spike tokens through SPS, then progressively encoded by $l$ stages of SpikeDS Blocks with 3D patch merging layers that double the channel dimensions , and finally classified via global average pooling and a linear classification head. (b) SpikeDS Block applies Dual Sparsity Spiking Attention (DSSA), 
which performs W-EMSA only on salient windows selected by firing-rate-guided pruning, 
followed by asymmetric Cross-Window Spiking Self-Attention (CW-SSA) for global 
context exchange, and a Spiking MLP. Window pruning is disabled in Stage~0 and 
activated with keep ratio~$\rho$ in Stages~$l$.}
\label{fig:overview}
\end{figure*}

Despite these advances, existing spiking transformers process all spatial tokens uniformly~\cite{liu2024sparsespikformer}, overlooking the inherent spatial redundancy in 3D medical volumes where large background regions carry negligible diagnostic information~\cite{mirab2025multi}. We introduce SpikeDS, a Dual Sparsity Spikformer that combines the activation sparsity inherent in SNNs with an explicit spatial sparsity mechanism. SpikeDS partitions the 3D volume into non-overlapping windows and prunes uninformative ones based on firing rates, concentrating attention computation on diagnostically relevant regions. A complementary cross-window attention module then maintains global information exchange between the retained and pruned windows through an asymmetric query–key scheme. This dual sparsity is applied progressively across encoder stages, allowing early layers to build reliable low-level features before deeper layers focus computation on the most informative regions.


\section{Method}

\subsection{Overall Architecture}

Fig~\ref{fig:overview} illustrates the overall architecture of our proposed method. A 3D input volume $\mathbf{X}\!\in\!\mathbb{R}^{C\times H\times W\times Z}$ is first encoded into spiking feature maps by a Spike Patch Splitting (SPS) module.
The tokens are then processed by a three-stage hierarchical encoder with increasing channel dimensions and decreasing spatial resolutions, connected via 3D patch merging layers.
Each stage stacks multiple SpikeDS Blocks that comprise a sequence of DSSA $\to$ CW-SSA $\to$ Spiking MLP, with residual connections throughout.
In Stage~0, all windows are processed without pruning; from Stage~1 and~2 onward, firing-rate-guided window pruning is activated.
Global average pooling, batch normalization, a LIF layer, and temporal mean reduction yield the final representation for classification.

\subsection{Spike Patch Splitting (SPS)}

SPS ~\cite{zhou2022spikformer} converts the continuous-valued input into discrete spike representations through four cascaded Conv3D$\,\to\,$BN$\,\to\,$LIF stages.
Spatial down-sampling via max pooling is applied selectively in the first two stages, reducing the resolution to one-quarter of the original.
A convolutional relative position encoding (RPE) branch injects local spatial priors before the tokens enter the transformer encoder.
We employ the Leaky Integrate-and-Fire (LIF) neuron model~\cite{burkitt2006review} with hard reset, where the membrane potential $U[t]$ accumulates input features and fires a binary spike $S[t]$ upon exceeding a threshold $V_\text{th}$:
\begin{equation}
U[t]=\alpha\!\cdot\!U[t\!-\!1]\!\cdot\!(1\!-\!S[t\!-\!1])+X[t],\;
S[t]=\Theta(U[t]-V_\text{th})
\label{eq:lif}
\end{equation}
where $\alpha=1-1/\tau$ is the membrane decay factor and $\Theta(\cdot)$ is the Heaviside step function.
Since the spike outputs are binary, downstream layers incur only accumulate (AC) operations rather than energy-intensive MAC operations.
Crucially, the first Conv3D layer of SPS receives continuous-valued inputs and therefore requires MAC operations; all subsequent layers operate in the spike domain and incur only AC operations.

\subsection{Dual Sparsity Spiking Attention (DSSA)}

DSSA unifies firing-rate-guided window pruning and window-based expert mixture spiking attention (W-EMSA) into a single module.
Given input $\mathbf{X}\!\in\!\mathbb{R}^{T\times B\times H\times W\times Z\times D}$, where $\mathbf{B}$ is the batch size, we partition it into $N_w$ non-overlapping 3D windows $\{W_j\}_{j=1}^{N_w}$ of size $w_H\!\times\!w_W\!\times\!w_Z$.
A saliency score for each window is computed as its mean firing rate:
\begin{equation}
s_j = \frac{1}{TBMD}\sum_{t,b,m,d} W_j^{(t,b,m,d)}
\label{eq:saliency}
\end{equation}
where $M\!=\!w_H w_W w_Z$ is the token count per window.
The active set $\mathcal{A}=\text{top-}k(\{s_j\},\,\lfloor\rho N_w\rfloor)$ is selected, where $\rho\!\in\!(0,1]$ denotes the proportion of windows retained after pruning,
and DSSA is defined as:
\begin{equation}
\text{DSSA}(W_j) =
\begin{cases}
\text{W-EMSA}(W_j), & j\in\mathcal{A}\\
W_j, & j\notin\mathcal{A}
\end{cases}
\label{eq:dssa}
\end{equation}
In Stage~0, pruning is disabled ($\mathcal{A}=\{1,\ldots,N_w\}$).

W-EMSA performs window-based expert mixture spiking attention inspired by SEMM~\cite{zhou2024spiking}.
Tokens pass through a LIF layer and are projected into queries, 
keys, and values via spiking linear layers 
(Linear$\,\to\,$BN$\,\to\,$LIF).
Since $\mathbf{Q}$, $\mathbf{K}$, $\mathbf{V}$ are all binary 
spike tensors after the LIF layers, the matrix products degenerate 
to sparse accumulate operations, yielding energy-efficient 
attention.
Each attention head is treated as a specialised expert:
\begin{equation}
\text{SSA}(\mathbf{Q},\mathbf{K},\mathbf{V}) 
= \text{SN}\!\bigl(
\mathbf{Q}\mathbf{K}^\top\mathbf{V}\cdot s\bigr),\;s = d^{-1/2}
\label{eq:attn}
\end{equation}

\begin{equation}
A_j^{(h)} = \text{SSA}^{(h)}(\mathbf{Q}_j^{(h)},\,
\mathbf{K}_j^{(h)},\,\mathbf{V}_j^{(h)})
\label{eq:expert}
\end{equation}
A spiking router generates a binary gating vector from the 
window-level mean token $\bar{w}_j$:
\begin{equation}
\mathbf{r}_j = \text{SN}\!\bigl(\text{BN}(\bar{w}_j\,
\mathbf{W}_R)\bigr) \in \{0,1\}^{H}
\label{eq:gate}
\end{equation}
The gated head outputs are concatenated and projected through 
a final spiking linear layer:
\begin{equation}
\mathbf{O}_j = \bigl[A_j^{(1)};\,\dots\,;\,A_j^{(H)}\bigr]
\in\{0,1\}^{M\times D}
\label{eq:head_concat}
\end{equation}
\begin{equation}
\text{W-EMSA}_j = \text{SN}\!\Bigl(\text{BN}\!\bigl(
(\mathbf{r}_j \odot \mathbf{O}_j)\,\mathbf{W}_O\bigr)\Bigr)
\label{eq:router}
\end{equation}
Since the router outputs binary spikes, each $r_j^{(h)} \in 
\{0,1\}$ acts as a binary mask that selectively activates or 
suppresses the corresponding head expert per window, preserving 
the spike-driven property of the network.

\begin{figure}[t]
\centering
\includegraphics[width=1.0\columnwidth]{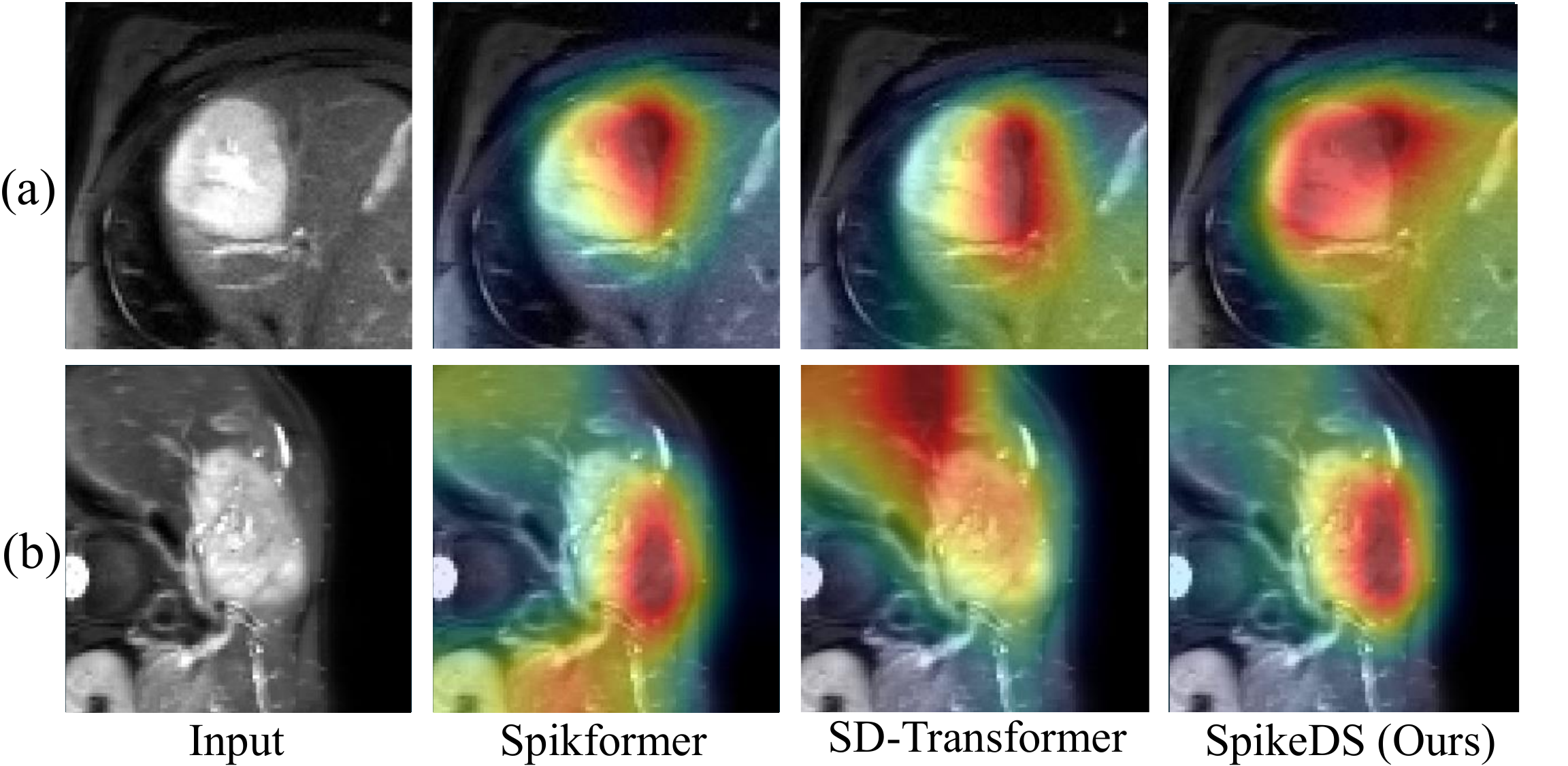}
\caption{Attention maps based on Spike Firing Rate (SFR). These visualizations illustrate the spatial density of spike activations within the input volume for (a) PNI-positive and (b) PNI-negative cases.}
\label{fig:gradcam}
\end{figure}

\subsection{Cross-Window Spiking Self-Attention (CW-SSA)}

CW-SSA enables global information exchange across all windows.
It is important to note that the window pruning introduced in DSSA excludes low-saliency windows only from the W-EMSA attention computation; the windows themselves and their feature representations are \emph{not} discarded.
CW-SSA exploits this by allowing pruned windows to remain accessible as key-value sources for global context aggregation.

Concretely, each window is mean-pooled to a representative token $\bar{\mathbf{w}}_j$.
When pruning is active, CW-SSA adopts an asymmetric query--key/value scheme:
\begin{equation}
  \mathbf{Q}\!=\!f_q(\{\bar{\mathbf{w}}_j\}_{j\in\mathcal{A}}),\;
  \mathbf{K}\!=\!f_k(\{\bar{\mathbf{w}}_j\}_{j=1}^{N_w}),\;
  \mathbf{V}\!=\!f_v(\{\bar{\mathbf{w}}_j\}_{j=1}^{N_w})
  \label{eq:cwssa}
\end{equation}
Queries are derived only from the $\rho N_w$ active windows, while keys and values are drawn from \emph{all} $N_w$ windows, including those pruned from W-EMSA.
This design is motivated by three considerations.
First, pruned windows retain their stage-input representations from the preceding patch merging layer (or SPS for Stage~0), which, although not refined by W-EMSA, still encode meaningful low-level spatial features.
Second, the overhead of projecting all $N_w$ tokens into keys and values is negligible because $N_w$ is the window-level token count, not the voxel-level count.
Third, the resulting attention map has shape $\rho N_w \times N_w$, allowing active windows to attend to the full spatial extent of the volume while the $\rho$ factor still reduces query-side computation.
The cross-attended tokens are broadcast back to active windows as a residual.
In Stage~0 (no pruning), CW-SSA operates symmetrically with $\rho\!=\!1$.

\begin{table}[t]
\centering
\caption{Performance and theoretical inference energy comparison for PNI prediction (5-fold CV). OPs denotes SOPs for SNNs and FLOPs for ANNs.}
\label{tab:main}
\setlength{\tabcolsep}{3.5pt}
\begin{tabular}{@{}lcccc@{}}
\toprule
Model & Param (M) & OPs (G) & Energy (mJ) & AUC \\
\midrule
ResNet-18~\cite{he2016deep}          & 33.2 & 9.16  & 42.2   & 0.670 \\
DenseNet-121~\cite{huang2017densely} & 11.3 & 11.5  & 52.7   & 0.711 \\
EfficientNet-B0~\cite{tan2019efficientnet} & 4.07 & 0.6 & 2.76 & 0.621 \\
\midrule
Spikformer~\cite{zhou2022spikformer}      & 4.29 & 8.40 & 16.0 & 0.703 \\
SD-Transformer~\cite{yao2023spike}      & 4.29 & 6.87 & 15.2 & 0.671 \\
\textbf{SpikeDS (Ours)}                   & 5.22   & 6.67   & 14.4   & 0.753    \\
\bottomrule
\end{tabular}
\end{table}

\subsection{SpikeDSBlock and Hierarchical Encoder}
 
Each SpikeDSBlock follows:
\begin{equation}
  \hat{\mathbf{X}} = \mathbf{X} + \text{DSSA}(\mathbf{X}) + \text{CW-SSA}(\text{DSSA}(\mathbf{X})),\;\;
  \label{eq:block}
\end{equation}
\begin{equation}
\mathbf{X}' = \hat{\mathbf{X}} + \text{SpikingMLP}(\hat{\mathbf{X}})
\end{equation}

The three-stage hierarchical encoder doubles the channel dimension and halves the spatial resolution at each transition via a 3D patch merging layer.
Window pruning is disabled in the first stage to allow reliable low-level feature formation, and is activated with keep ratio~$\rho$ in the deeper stages.
\section{Experiments}

\subsection{Dataset and Implementation Details}

\textbf{Dataset.}
We utilised a retrospective clinical cohort of 139 CCA patients (PNI-positive: 85, PNI-negative: 54) with pathologically confirmed PNI status, collected over more than a decade at Samsung Medical Center.
The T2-weighted phase of contrast-enhanced MRI was used as the single-phase input.
Each volume was cropped to a fixed $96\!\times\!96\!\times\!48$ region centred on the tumor~\cite{liu2024noninvasive}.
A three-channel input was constructed by stacking the normalized intensity image, a liver mask, and a tumor mask.

\textbf{Implementation.}
SpikeDS was implemented in PyTorch with SpikingJelly~\cite{fang2023spikingjelly} and evaluated using stratified 5-fold cross-validation.
The model was optimized with AdamW~\cite{loshchilov2019decoupled} (learning rate $1\!\times\!10^{-4}$, weight decay $1\!\times\!10^{-5}$) for 100 epochs with a batch size of~2 using mixed-precision training on a single NVIDIA~A100 GPU.
A cosine annealing schedule was applied.
The base channel dimension was set to $C\!=\!64$ and the number of LIF timesteps to $T\!=\!4$.
\subsection{Quantitative Evaluation and Visualization}

We benchmarked SpikeDS against CNN and SNN baselines using the identical three-channel input (intensity, liver mask, tumor mask). As shown in Table~\ref{tab:main}, 3D CNN models achieved AUCs between 0.621 and 0.711, while SNN baselines attained comparable AUCs at substantially lower energy consumption. SpikeDS achieves the highest AUC of 0.753 at only 14.4\,mJ,
the lowest among all SNN baselines.
Although EfficientNet-B0 consumes less energy (2.76\,mJ),
its AUC of 0.621 is the lowest in the comparison,
whereas SpikeDS maintains the best AUC--energy trade-off
across all models.

Fig.~\ref{fig:gradcam} visualizes the spiking firing rate (SFR) maps on representative axial slices. SpikeDS concentrates higher firing rates around the tumor region, indicating that the dual sparsity mechanism effectively directs computational focus toward diagnostically relevant areas.



\subsection{Ablation Study}
Table~\ref{tab:ablation} summarizes the ablation results.
Removing window pruning raises energy to 16.17\,mJ and lowers AUC to 0.674, as indiscriminate full-window attention introduces background noise.
Disabling CW-SSA reduces AUC to 0.732. Replacing its asymmetric design with a symmetric variant that excludes pruned windows from the key–value pool further confirms the value of pruned-window context: the asymmetric design achieves 0.753 AUC, while the symmetric variant drops to 0.737.
Replacing expert routing with standard W-SSA drops AUC to 0.729 at 16.16\,mJ, losing head-wise suppression of uninformative attention.
Pruning at all stages degrades AUC to 0.675 because early features are insufficiently formed for reliable saliency estimation, while pruning only at Stage~2 yields suboptimal efficiency of 14.47\,mJ.
The adopted configuration---pruning with $\rho\!=\!0.5$ at Stages~1 and~2 only---achieves the best trade-off, achieving AUC of 0.753 at 14.41\,mJ.

\begin{table}[t]
\centering
\caption{Ablation study on the core modules of SpikeDS.}
\label{tab:ablation}
\setlength{\tabcolsep}{4pt}
\begin{tabular}{@{}lcc@{}}
\toprule
Configuration & AUC & Energy (mJ) \\
\midrule
w/o Window Pruning (all stages full)    & 0.674& 16.17 \\
w/o CW-SSA                              & 0.732  & 14.20 \\
CW-SSA symmetric (Q,K,V from active)    & 0.737  & 14.46 \\
w/o Expert Routing (W-EMSA$\to$W-SSA)   & 0.729  & 16.16\\
Pruning at Stage~2 only                 & 0.737  & 14.47 \\
Pruning at all stages                   & 0.675  & 14.10 \\
\textbf{SpikeDS (Ours)}                 & 0.753  & 14.41 \\
\bottomrule
\end{tabular}
\end{table}

\subsection{Energy Analysis}
SpikeDS combines \emph{activation sparsity} (binary spikes) and \emph{spatial sparsity} (window pruning). All energy estimates adopt the 45\,nm model~\cite{horowitz20141} with $E_\text{MAC}=4.6$\,pJ and $E_\text{AC}=0.9$\,pJ. Binary spike activations reduce MACs to ACs in all layers except the first (float-input) Conv3D, following the standard SOP formulation~\cite{zhou2022spikformer,yao2023spike}:
\begin{equation}
\text{SOP}(l) = \bar{s}_l \cdot T \cdot \text{FLOPs}(l)
\label{eq:sop}
\end{equation}
Each SpikeDS Block incurs energy modulated by the keep ratio $\rho$:
\begin{equation}
E_\text{block} = E_\text{AC}\cdot\bigl(\rho\,\Phi_\text{E} + \Phi_\text{C} + \Phi_\text{M}\bigr)
\label{eq:block_energy}
\end{equation}
where $\Phi_\text{E}$, $\Phi_\text{C}$, and $\Phi_\text{M}$ denote the SOP counts of W-EMSA, CW-SSA, and Spiking MLP, respectively. When $\rho\!=\!1$ (Stage~0) the block reduces to a standard spiking transformer; when $\rho\!<\!1$ (Stages~1--2), W-EMSA cost decreases proportionally.

Total inference energy is:
\begin{multline}
E_\text{total} = E_\text{MAC}\cdot\text{FL}_1 \\
+ E_\text{AC}\cdot\Bigl(\text{SOP}_\text{SPS} + \sum_{i}\bigl[d_i\cdot\tfrac{E_\text{block}^{(i)}}{E_\text{AC}} + \text{SOP}_\text{PM}^{(i)}\bigr] + \text{SOP}_\text{head}\Bigr)
\label{eq:total_energy}
\end{multline}
where $d_i$ is the block depth of stage $i$ ($d=2,2,4$).

Activation sparsity alone ($\bar{s}\approx0.10$) yields ${\sim}82\%$ energy saving over the ANN-equivalent configuration of the same architecture (92.3\,mJ $\to$ 16.17\,mJ). Window pruning at Stages~1--2 ($\rho=0.5$) compounds this further for a total saving of 84\% (92.3\,mJ $\to$ 14.41\,mJ).

\section{Conclusion}

We introduced SpikeDS, a spiking transformer that exploits activation sparsity and spatial sparsity for energy-efficient PNI prediction in 3D MRI.
By introducing DSSA, which unifies firing-rate window pruning with W-EMSA, and CW-SSA, which preserves global context even for pruned regions through asymmetric cross-window attention, SpikeDS achieves a principled balance between computational efficiency and classification performance.
The progressive pruning strategy ensures that spatial sparsity is applied only where saliency estimation is reliable.
We provided a decomposed energy analysis that explicitly separates the contributions of activation sparsity and spatial sparsity, demonstrating that their compounding effect yields substantial savings.
Evaluated on a clinical CCA cohort for PNI prediction, SpikeDS demonstrates the potential of dual-sparsity spiking architectures for volumetric medical imaging under stringent energy constraints.

\section*{Acknowledgment}
This work was supported by the IITP grant (IITP-2023-RS-2023-00256081) funded by MSIT, Korea, and the ANCHOR program (2026-ANCHOR-01-110) funded by the Ministry of Education and the Seoul Metropolitan Government, Republic of Korea.

\bibliographystyle{IEEEtran}
\bibliography{references}

@article{liebig2009perineural,
  title={Perineural invasion in cancer: a review of the literature},
  author={Liebig, Catherine and Ayala, Gustavo and Wilks, Jonathan A and Berger, David H and Albo, Daniel},
  journal={Cancer},
  volume={115},
  number={15},
  pages={3379--3391},
  year={2009},
  publisher={Wiley}
}

@article{conti2026perineural,
  title={Perineural invasion is a prognostic factor in cholangiocarcinoma, regardless of anatomical location: a systematic review and meta-analysis},
  author={Conti, Sefora and Tissera, Natalia S and Castet, Florian and others},
  journal={JHEP Reports},
  pages={101770},
  year={2026},
  publisher={Elsevier}
}

@article{li2020perineural,
  title={Perineural invasion of hilar cholangiocarcinoma in {C}hinese population: one center's experience},
  author={Li, Cheng-Gang and Zhou, Zhi-Peng and Tan, Xiang-Long and Zhao, Zhi-Ming},
  journal={World Journal of Gastrointestinal Oncology},
  volume={12},
  number={4},
  pages={457},
  year={2020}
}

@article{wei2022prognostic,
  title={Prognostic impact of perineural invasion in intrahepatic cholangiocarcinoma: multicentre study},
  author={Wei, Tao and Zhang, Xu-Feng and He, Jin and others},
  journal={British Journal of Surgery},
  volume={109},
  number={7},
  pages={610--616},
  year={2022},
  publisher={Oxford University Press}
}

@article{huang2021feasibility,
  title={Feasibility of magnetic resonance imaging-based radiomics features for preoperative prediction of extrahepatic cholangiocarcinoma stage},
  author={Huang, Xiao and Cheng, Jian and Zhang, Lei and others},
  journal={European Journal of Cancer},
  volume={155},
  pages={227--235},
  year={2021}
}

@article{larue2017quantitative,
  title={Quantitative radiomics studies for tissue characterization: a review of technology and methodological procedures},
  author={Larue, Ruben THM and Defraene, Gilles and De Ruysscher, Dirk and Lambin, Philippe and Van Elmpt, Wouter},
  journal={The British Journal of Radiology},
  volume={90},
  number={1070},
  pages={20160665},
  year={2017}
}

@article{lambin2017radiomics,
  title={Radiomics: the bridge between medical imaging and personalized medicine},
  author={Lambin, Philippe and Leijenaar, Ralph TH and Deist, Timo M and others},
  journal={Nature Reviews Clinical Oncology},
  volume={14},
  number={12},
  pages={749--762},
  year={2017}
}

@inproceedings{he2016deep,
  title={Deep residual learning for image recognition},
  author={He, Kaiming and Zhang, Xiangyu and Ren, Shaoqing and Sun, Jian},
  booktitle={Proc. IEEE/CVF CVPR},
  pages={770--778},
  year={2016}
}

@inproceedings{huang2017densely,
  title={Densely connected convolutional networks},
  author={Huang, Gao and Liu, Zhuang and Van Der Maaten, Laurens and Weinberger, Kilian Q},
  booktitle={Proc. IEEE/CVF CVPR},
  pages={4700--4708},
  year={2017}
}

@inproceedings{tan2019efficientnet,
  title={{EfficientNet}: Rethinking model scaling for convolutional neural networks},
  author={Tan, Mingxing and Le, Quoc},
  booktitle={Proc. ICML},
  pages={6105--6114},
  year={2019}
}

@inproceedings{liu2021swin,
  title={Swin {T}ransformer: Hierarchical vision transformer using shifted windows},
  author={Liu, Ze and Lin, Yutong and Cao, Yue and Hu, Han and Wei, Yixuan and Zhang, Zheng and Lin, Stephen and Guo, Baining},
  booktitle={Proc. IEEE/CVF ICCV},
  pages={10012--10022},
  year={2021}
}

@article{roy2019towards,
  title={Towards spike-based machine intelligence with neuromorphic computing},
  author={Roy, Kaushik and Jaiswal, Akhilesh and Panda, Priyadarshini},
  journal={Nature},
  volume={575},
  number={7784},
  pages={607--617},
  year={2019}
}

@article{yin2021accurate,
  title={Accurate and efficient time-domain classification with adaptive spiking recurrent neural networks},
  author={Yin, Bojian and Corradi, Federico and Boht{\'e}, Sander M},
  journal={Nature Machine Intelligence},
  volume={3},
  number={10},
  pages={905--913},
  year={2021}
}

@inproceedings{zhou2022spikformer,
  title={Spikformer: When spiking neural network meets transformer},
  author={Zhou, Zhaokun and Zhu, Yuesheng and He, Chao and Wang, Yaowei and Yan, Shuicheng and Tian, Yonghong and Yuan, Li},
  booktitle={Proc. ICLR},
  year={2023}
}

@article{yao2023spike,
  title={Spike-driven transformer},
  author={Yao, Man and Hu, Jiakui and Zhou, Zhaokun and Yuan, Li and Tian, Yonghong and Xu, Bo and Li, Guoqi},
  journal={Advances in Neural Information Processing Systems},
  volume={36},
  pages={64043--64058},
  year={2023}
}

@article{burkitt2006review,
  title={A review of the integrate-and-fire neuron model: {I}. {H}omogeneous synaptic input},
  author={Burkitt, Anthony N},
  journal={Biological Cybernetics},
  volume={95},
  number={1},
  pages={1--19},
  year={2006}
}

@article{fang2023spikingjelly,
  title={{SpikingJelly}: An open-source machine learning infrastructure platform for spike-based intelligence},
  author={Fang, Wei and Chen, Yanqi and Ding, Jianhao and Yu, Zhaofei and Masquelier, Timoth{\'e}e and Chen, Ding and Huang, Tiejun and Tian, Yonghong},
  journal={Science Advances},
  volume={9},
  number={40},
  pages={eadi1480},
  year={2023}
}

@inproceedings{loshchilov2019decoupled,
  title={Decoupled weight decay regularization},
  author={Loshchilov, Ilya and Hutter, Frank},
  booktitle={Proc. ICLR},
  year={2019}
}

@article{liu2024noninvasive,
  title={Noninvasive prediction of perineural invasion in intrahepatic cholangiocarcinoma with interpretable machine learning based on {MRI}},
  author={Liu, Ziwei and Luo, Chun and Chen, Xinjie and others},
  journal={International Journal of Surgery},
  volume={110},
  number={2},
  pages={1039--1051},
  year={2024}
}

@inproceedings{liu2024sparsespikformer,
  title={Sparsespikformer: A co-design framework for token and weight pruning in spiking transformer},
  author={Liu, Yue and Xiao, Shanlin and Li, Bo and Yu, Zhiyi},
  booktitle={ICASSP 2024-2024 IEEE International Conference on Acoustics, Speech and Signal Processing (ICASSP)},
  pages={6410--6414},
  year={2024},
  organization={IEEE}
}

@article{mirab2025multi,
  title={A multi-scale attention-based Swin transformer model for medical images segmentation},
  author={Mirab Golkhatmi, Benyamin and Houshmand, Mahboobeh and Hosseini, Seyyed Abed},
  journal={Scientific Reports},
  volume={15},
  number={1},
  pages={38893},
  year={2025},
  publisher={Nature Publishing Group UK London}
}

@article{yao2024spike,
  title={Spike-driven transformer v2: Meta spiking neural network architecture inspiring the design of next-generation neuromorphic chips},
  author={Yao, Man and Hu, JiaKui and Hu, Tianxiang and Xu, Yifan and Zhou, Zhaokun and Tian, Yonghong and Xu, Bo and Li, Guoqi},
  journal={arXiv preprint arXiv:2404.03663},
  year={2024}
}

@article{zhou2024spiking,
  title={Spiking transformer with experts mixture},
  author={Zhou, Zhaokun and Lu, Yijie and Jia, Yanhao and Che, Kaiwei and Niu, Jun and Huang, Liwei and Shi, Xinyu and Zhu, Yuesheng and Li, Guoqi and Yu, Zhaofei and others},
  journal={Advances in Neural Information Processing Systems},
  volume={37},
  pages={10036--10059},
  year={2024}
}

@inproceedings{horowitz20141,
  title={1.1 computing's energy problem (and what we can do about it)},
  author={Horowitz, Mark},
  booktitle={2014 IEEE international solid-state circuits conference digest of technical papers (ISSCC)},
  pages={10--14},
  year={2014},
  organization={IEEE}
}

@article{meng2023clinical,
  title={The clinical implications and molecular features of intrahepatic cholangiocarcinoma with perineural invasion},
  author={Meng, Xian-Long and Lu, Jia-Cheng and Zeng, Hai-Ying and Chen, Zhen and Guo, Xiao-Jun and Gao, Chao and Pei, Yan-Zi and Hu, Shu-Yang and Ye, Mu and Sun, Qi-Man and Yang, Guo-Huang and Cai, Jia-Bin and Huang, Pei-Xin and Yv, Lei and Zhang, Lv and Shi, Ying-Hong and Ke, Ai-Wu and Zhou, Jian and Fan, Jia and Chen, Yi and Huang, Xiao-Yong and Shi, Guo-Ming},
  journal={Hepatology International},
  volume={17},
  number={1},
  pages={63--76},
  year={2023},
  publisher={Springer}
}

@article{chen2025preoperative,
  title={Preoperative MRI prediction and molecular pathway study of perineural invasion in intrahepatic cholangiocarcinoma: Insights from bioinformatics approach},
  author={Chen, Mei-Cheng and Zhou, Xiao-Qi and Liu, Zi-Wei and Tang, Mi-Mi and Chen, Yu-Ying and Luo, Yan-Ji and Ma, Ling and Liao, Bing and Feng, Shi-Ting},
  journal={European Journal of Surgical Oncology},
  pages={110546},
  year={2025},
  publisher={Elsevier}
}

@inproceedings{han2026losanet,
  title={{LoSA-Net}: A Localized and Scale-Adaptive Network for Boundary-Sensitive Prediction of Perineural Invasion in 3D MRI},
  author={Han, Youngung and Go, Hyunsu and Kim, Kyeonghun and Um, Induk and Kim, Junga and Jung, Jaewon and Kim, Nam-Joon and Jeong, Woo Kyoung and Lee, Won Jae and Liao, Ken Ying-Kai and Hong, Pa and Lee, Hyuk-Jae},
  booktitle={2026 IEEE 23rd International Symposium on Biomedical Imaging (ISBI)},
  pages={1--5},
  year={2026},
  organization={IEEE},
  doi={10.1109/ISBI61048.2026.11515516}
}

@inproceedings{han2026mmaformer,
  title={{MMA-Former}: Multi-Window Mixture-of-Head Attention Transformer for Adaptive PNI Prediction in 3D MRI},
  author={Han, Youngung and Um, Induk and Kim, Kyeonghun and Kim, Junga and Go, Hyunsu and Jung, Jaewon and Kim, Nam-Joon and Jeong, Woo Kyoung and Lee, Won Jae and Hong, Pa and Liao, Ken Ying-Kai and Lee, Hyuk-Jae},
  booktitle={2026 IEEE 23rd International Symposium on Biomedical Imaging (ISBI)},
  pages={1--5},
  year={2026},
  organization={IEEE},
  doi={10.1109/ISBI61048.2026.11515401}
}

@article{han2026neonet,
  title={{NeoNet}: An End-to-End 3D MRI-Based Deep Learning Framework for Non-Invasive Prediction of Perineural Invasion via Generation-Driven Classification},
  author={Han, Youngung and Cha, Minkyung and Kim, Kyeonghun and Um, Induk and Sho, Myeongbin and Bae, Joo Young and Jung, Jaewon and Park, Jung Hyeok and Lee, Seojun and Kim, Nam-Joon and Jeong, Woo Kyoung and Lee, Won Jae and Hong, Pa and Liao, Ken Ying-Kai and Lee, Hyuk-Jae},
  journal={arXiv preprint arXiv:2603.29449},
  year={2026}
}

\end{document}